\documentclass[sigconf, natbib=false]{acmart}
\usepackage{graphicx}
\usepackage{subcaption}
\usepackage{booktabs}
\usepackage{multirow}
\usepackage{tabularx}
\usepackage{caption}
\usepackage{adjustbox}
\AtBeginDocument{%
  }

\setcopyright{acmlicensed}
\copyrightyear{2026}
\acmYear{2026}
\acmDOI{XXXXXXX.XXXXXXX}
\acmConference[Conference acronym 'XX]{Make sure to enter the correct
  conference title from your rights confirmation email}{June 03--05,
  2018}{Woodstock, NY}
\acmISBN{978-1-4503-XXXX-X/2018/06}

\RequirePackage[
  datamodel=acmdatamodel,
  style=acmnumeric,
  ]{biblatex}

\begin{document}

\title{BRIDGE: Benchmark for multi-hop Reasoning In long multimodal Documents with Grounded Evidence}


\author{Biao Xiang}
\affiliation{%
  \institution{The University of Melbourne}
  \city{Melbourne}
  \country{Australia}}
\email{biaox@student.unimelb.edu.au}

\author{Soyeon Caren Han}
\affiliation{%
  \institution{The University of Melbourne}
  \city{Melbourne}
  \country{Australia}}
\email{Caren.Han@unimelb.edu.au}

\author{Yihao Ding}
\affiliation{%
  \institution{The University of Western Australia}
  \city{Perth}
  \country{Australia}}
\email{yihao.ding@uwa.edu.au}


\begin{abstract}
Multi-hop question answering (QA) is widely used to evaluate the reasoning capabilities of large language models, yet most benchmarks focus on final answer correctness and overlook intermediate reasoning, especially in long multimodal documents. 
We introduce BRIDGE, a benchmark for multi-hop reasoning over long scientific papers that require integrating evidence across text, tables, and figures. The dataset supports both chain-like and fan-out structures and provides explicit multi-hop reasoning annotations for step-level evaluation beyond answer accuracy. 
Experiments with state-of-the-art LLMs and multimodal retrieval-augmented generation (RAG) systems reveal systematic deficiencies in evidence aggregation and grounding that remain hidden under conventional answer-only evaluation. BRIDGE provides a targeted testbed for diagnosing reasoning failures in long multimodal documents.
\end{abstract}



\received{20 February 2007}
\received[revised]{12 March 2009}
\received[accepted]{5 June 2009}

\maketitle

\section{Introduction}
Recent advances in Large Language Models (LLMs) have significantly improved document-based question answering (QA). However, in high-stakes domains such as finance \cite{ding2023formnludatasetformnatural,TITO2023109834,mathew2021docvqadatasetvqadocument}, healthcare \cite{soni-etal-2022-radqa,jin-etal-2019-pubmedqa}, and academic research \cite{pramanick2025spiqadatasetmultimodalquestion}, answers are rarely explicitly stated and must be derived through multi-hop reasoning over heterogeneous evidence distributed across long documents.
Existing multi-hop QA benchmarks are typically framed as either fan-out or chain-like \cite{zhu2024fanoutqamultihopmultidocumentquestion}, differing in how reasoning structure is enforced and evaluated. Fan-out formulations emphasize parallel evidence collection with answer-level aggregation, often without constraining intermediate reasoning. In contrast, chain-like formulations impose sequential dependency, where errors propagate along the reasoning path and shortcut behaviors can be exposed \cite{chen-durrett-2019-understanding}. Nevertheless, as summarized in Table~\ref{tab:dataset_comparison}, most benchmarks lack explicit supervision and step-level evaluation of intermediate reasoning \cite{wu2024cofcastepwisecounterfactualmultihop}. 
This limitation becomes more pronounced in multimodal settings. Although several datasets incorporate multiple modalities \cite{wu2025mmqa,nararatwong-etal-2024-dbqr,ishii-etal-2024-jemhopqa,tan2024robusttemporalreasoninglarge,kim2025biohoprbenchmarkmultihopmultianswer}, modalities are frequently treated as independent or redundant information sources. As a result, models can rely primarily on textual cues while underutilizing tables or figures, reducing multimodal reasoning to shallow pattern matching rather than structured evidence aggregation.
We argue that long scientific papers provide a natural and challenging testbed for structured multi-hop multimodal reasoning. Unlike loosely connected Wikipedia passages \cite{yang2018hotpotqadatasetdiverseexplainable,2wikimhqa}, scientific documents exhibit coherent discourse structures in which claims introduced in text are quantified in tables and validated in figures, forming intrinsic cross-modal dependency chains. Evaluating reasoning in such settings requires not only answer correctness, but also verification of intermediate evidence usage and cross-modal consistency.

\begin{table}[t] 
\centering
\caption{Comparison with existing multi-hop QA benchmarks. MH: Multi-hop; Type: Question types (Ab: Abstractive; Cp: Comparative; Re: Causal Reasoning); Ann.: Explicit multi-hop annotation; Evl.: Step-by-step multi-hop evaluation; MM: Multimodal evidence.}
\vspace{-1.5em}
\label{tab:dataset_comparison}
\footnotesize 
\setlength{\tabcolsep}{2pt} 
\begin{tabularx}{\columnwidth}{l c l c c c X r} 
\toprule
\textbf{Dataset} & \textbf{MH} & \textbf{Type} & \textbf{Ann.} & \textbf{Evl.} & \textbf{MM} & \textbf{Doc. Scope} & \textbf{Scale} \\
\midrule
MultiTabQA \cite{pal-etal-2023-multitabqa} & $\checkmark$ & Cp & $\times$ & $\times$ & $\times$ & Relational DB & 9k \\
MEQA \cite{NEURIPS2024_meqa}      & $\checkmark$ & Cp+Re & $\times$ & $\times$ & $\times$ & Multi-doc text & 2,243 \\
SPIQA  \cite{pramanick2025spiqadatasetmultimodalquestion}    & $\times$     & ---   & $\times$ & $\times$ & $\checkmark$ & Sci. papers & 270k \\
DBQR-QA \cite{nararatwong-etal-2024-dbqr}   & $\checkmark$ & Cp+Re & $\times$ & $\times$ & $\checkmark$ & Finance reports & 400 \\
JEMHopQA \cite{ishii-etal-2024-jemhopqa}   & $\checkmark$ & Cp+Re & $\checkmark$ & $\times$ & $\times$ & Wikipedia (JP) & 1179 \\
Complex-TR \cite{tan2024robusttemporalreasoninglarge} & $\checkmark$ & Cp+Re & $\times$ & $\times$ & $\times$ & Temporal facts & 60k \\
FanOutQA \cite{zhu2024fanoutqamultihopmultidocumentquestion}   & $\checkmark$ & Cp & $\times$ & $\times$ & $\times$ & Wikipedia & 1894 \\
MultiHoax \cite{shafiei2025multihoaxdatasetmultihopfalsepremise}  & $\checkmark$ & Cp+Re & $\times$ & $\times$ & $\times$ & Wikipedia & 6,944 \\
BioHopR \cite{kim2025biohoprbenchmarkmultihopmultianswer}    & $\checkmark$ & Re & $\times$ & $\times$ & $\times$ & Biomed. KG & 10k \\
Cofca  \cite{wu2024cofcastepwisecounterfactualmultihop}    & $\checkmark$ & Re & $\checkmark$ & $\checkmark$ & $\times$ & Short passages & 3.6k \\
MMQA  \cite{wu2025mmqa}     & $\checkmark$ & Cp & $\times$ & $\times$ & $\checkmark$ & Relational tables & 3.3k \\
\midrule
\textbf{BRIDGE (Ours)}      & $\checkmark$ & Ab+Cp+Re & $\checkmark$ & $\checkmark$ & $\checkmark$ & Sci. papers & 11k \\
\bottomrule
\end{tabularx}
\vspace{-1.5em}
\end{table}

Our main contributions can be summarized as: (1). We introduce a \textbf{B}enchmark for multi-hop \textbf{R}easoning \textbf{I}n long multimodal \textbf{D}ocuments with \textbf{G}rounded
\textbf{E}vidence (BRIDGE), that supports both chain-like and fan-out question formulations. The benchmark covers diverse question types, including comparative, causal reasoning, and abstractive questions.
(2). We construct a multimodal multi-hop QA benchmark grounded in long scientific documents, requiring explicit reasoning across text, tables, and figures within long documents. 
(3). We provide explicit multi-hop reasoning annotations and evaluation protocols that assess intermediate reasoning states and evidence usage, going beyond final answer correctness. In addition, we introduce a structured error taxonomy to facilitate fine-grained analysis of reasoning failures.

\section{Dataset}
We present the \textsc{BRIDGE} dataset, supporting both fan-out and chain-like multi-hop QA over long multimodal scientific documents.

\begin{figure}[t]
\centering
\small
\setlength{\tabcolsep}{3pt}
\renewcommand{\arraystretch}{1.1}

\begin{tabularx}{\columnwidth}{@{}c X X@{}}
\toprule
\textbf{Type} & \textbf{Questions (and Mod.)} & \textbf{Answers (shorter version)} \\
\midrule
Cp & In Fig 3, which task sets a larger ``Ann'' size, task 1 or 3? (F+T)
& They are the same. Both of them are 110.
\\
Ab & Based on the whole paper, how do explanation types shift with subjectivity? (T)
& Justification dominates, while argument increases as tasks become more subjective.
\\
Re & Why use only a subset of fallacy types? (T+Tb)
& To reduce label ambiguity, and make annotation and evaluation reliable in Table~2.
\\
\bottomrule
\end{tabularx}

\vspace{0.5em}

\includegraphics[width=\columnwidth]{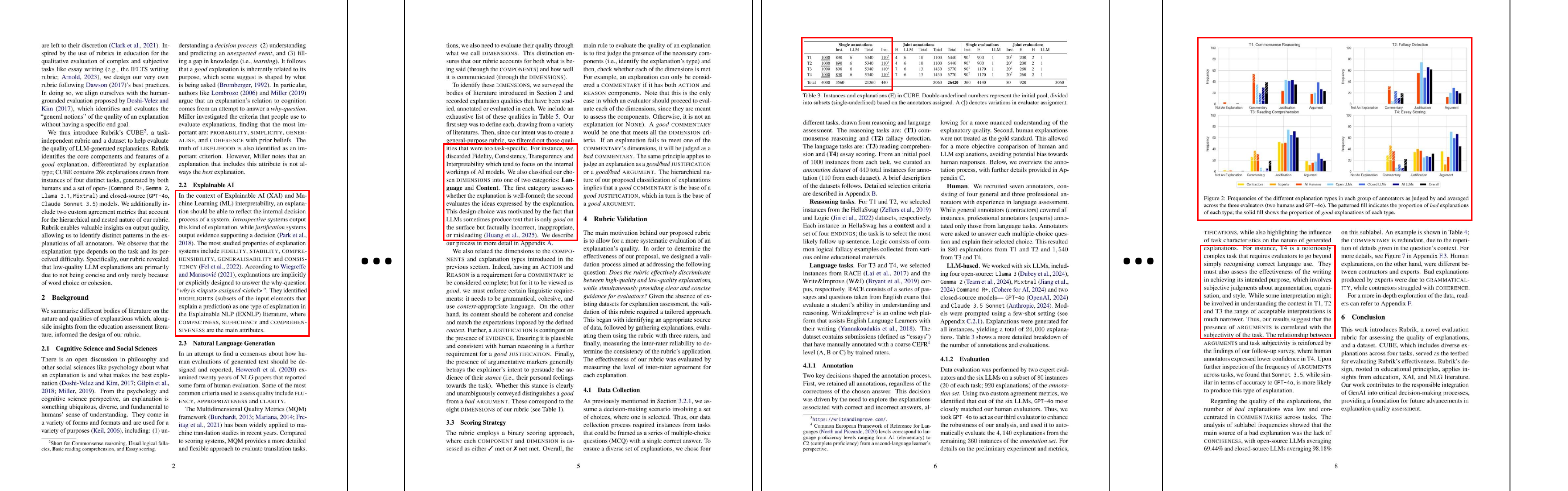}

\caption{Representative examples of comparative (Cp), abstractive (Ab), and causal reasoning (Re) questions (top), and the corresponding pages where evidences locate (bottom). Mod.: involved modalities (T: text; Tb: table; F: figure).}
\vspace{-2em}
\label{fig:examples_and_pages}
\end{figure}

\noindent\textbf{Task Definition.}
The \textsc{BRIDGE} task is formally defined as follows: given a question $q$ and a multi-page scientific document $D$—a long-form scientific paper comprising heterogeneous and interdependent modalities such as textual passages, tables, and figures—the goal is to generate a final answer $a$ along with a set of supporting evidences $E = \{e_1, e_2, \ldots, e_k\}$. The document is represented as a collection of semantic entities, each associated with a specific location (e.g., page index and region) and consisting of text spans, table entries, or visual figures. The objective is to require the model $\mathcal{F}$ to perform grounded reasoning over $D$, such that $\mathcal{F}(q, D) \rightarrow (a, E)$. The task typically demands \textit{multi-hop reasoning}, where answering $q$ necessitates synthesizing information from multiple evidences, potentially across modalities. Two types of reasoning structures are considered: \textit{chain-like}, where evidences must be followed in a sequential, dependent manner; and \textit{fan-out}, where multiple evidences contribute in parallel to the final answer. This formulation allows for comprehensive evaluation of a model's capacity to perform complex, modality-spanning reasoning within scientific documents.

\noindent\textbf{Data Collection and Preprocessing}
We construct the dataset from 262 research papers on ArXiv\footnote{https://info.arxiv.org/help/api/index.html}, including PDF files and available LaTeX sources. To ensure structural quality, we focus on papers published at top-tier venues. The corpus spans recent computer science research (2023–2025), primarily in NLP and computer vision (e.g., ACL, EMNLP, CVPR, ICCV). Each paper is parsed using the Adobe PDF Extract API\footnote{https://developer.adobe.com/document-services/apis/pdf-extract/} to extract text, tables, and figures with page indices and bounding-box metadata, enabling layout-aware document representations.


\subsection{QA Pair Generation}
We generate multi-hop question–answer pairs using a chain-of-thought (CoT) prompting strategy with LLMs.

\noindent\textbf{Defining Question Types}  We define three types of multi-hop questions based on the underlying reasoning patterns:
(1) \textit{Causal-reasoning questions}, where hops are connected through causal relations between entities or events;
(2) \textit{Comparative questions}, where multiple hops involve comparisons across entities, primarily numerical values, and occasionally concepts;
(3) \textit{Abstractive questions}, which require a holistic understanding of the entire paper to produce a summary-style answer.

\noindent\textbf{Question Type Guided Prompt Designing} 
We adopt a two-stage prompting framework: Stage 1 (Structure Mining) extracts question-type-specific entity structures, and Stage 2 (Constraint-Guided Generation), conditions on these structures, formal question definitions, and multi-hop constraints to generate structured multi-hop QA pairs with explicit evidence hops. 

\noindent\textbf{Dual-Stage QA Quality Filtering}
To ensure high data reliability, we employ a two-stage filtering pipeline: (1) a rule-based pre-filter that eliminates malformed instances, unresolved references, and instances lacking numeric grounding or metadata; and (2) an LLM-as-a-judge framework that evaluates grounding, faithfulness, and reasoning depth. By utilizing structured PDF anchors to flag hallucinations and "single-hop shortcuts," this process ensures a scalable, reproducible collection of high-confidence multi-hop QA pairs.

\subsection{Dataset Analysis}

\begin{figure}[t]
  \centering
  \includegraphics[width=\columnwidth]{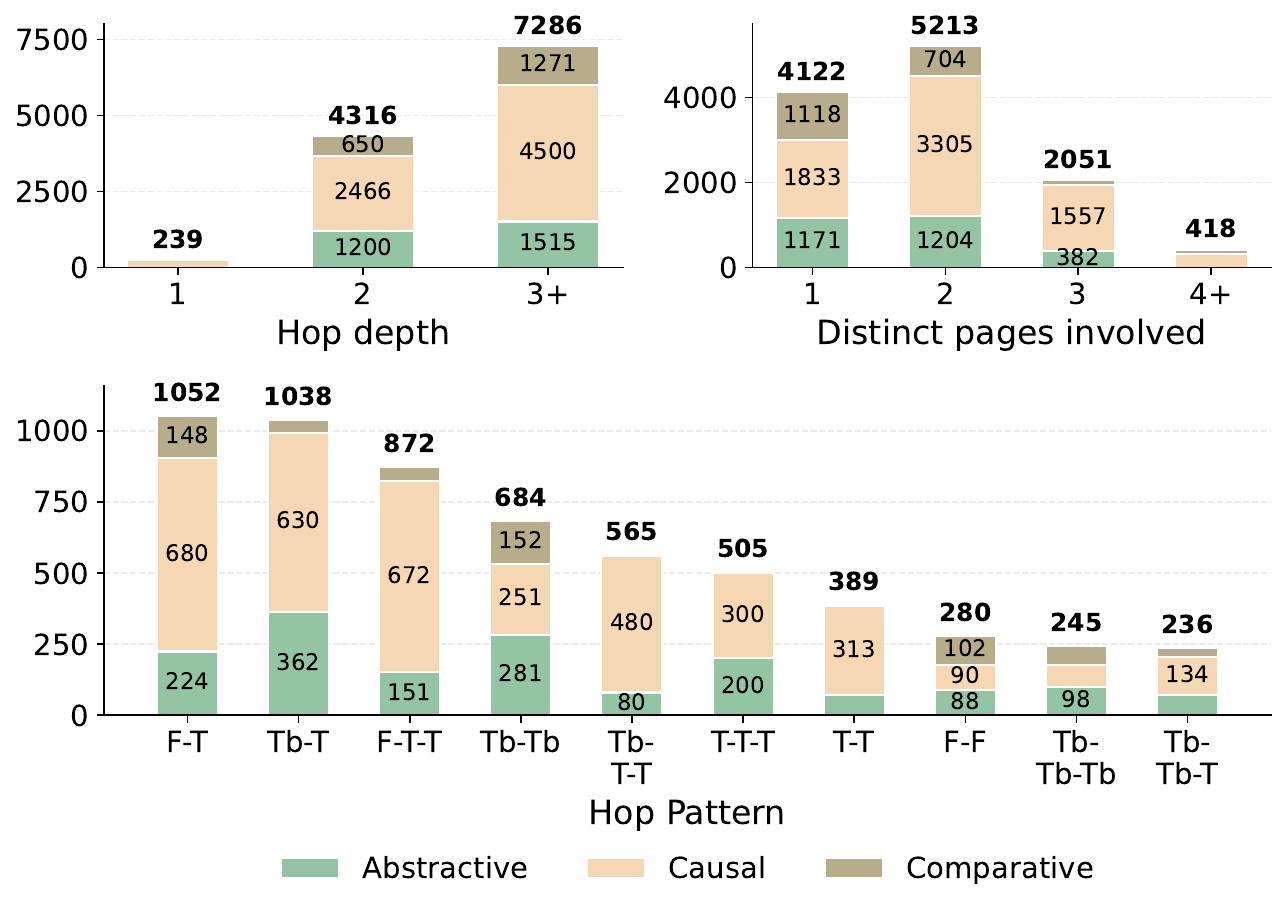}
  \caption{Distribution of QA instances by hop depth, number of distinct pages involved, and hop pattern, broken down by question type (Abstractive, Causal, Comparative)}
  \label{fig:qadis2}
  \vspace{-0.5em}
\end{figure}
BRIDGE contains 11,857 QA pairs with annotated evidence chains, spanning three task types and diverse hop patterns (e.g., table–text, figure–table–text). We summarize its structural properties along three dimensions.
\textit{i) Hop Depth.} 
Most questions require two reasoning steps, with a substantial portion involving three or more hops. It indicates a broader range of reasoning depths.
\textit{ii) Cross-Page Scope.} 
A large proportion of questions aggregate evidence from multiple pages, commonly two to three and extending beyond, reflecting the long-document nature.
\textit{iii) Modality Transitions.} 
We observe diverse modality transitions, including text–table, text–figure, and multi-step cross-modal chains. While purely unimodal reasoning forms only a subset.

\section{Experimental Setup}
We evaluate representative MLLMs, including closed-source systems (Gemini-3 \cite{google_gemini}, ChatGPT-5\cite{openai_chatgpt}) and open-weight models (Qwen3 \cite{qwen_tech}, Gemma-3 \cite{google_gemma}), under a unified evaluation pipeline. All models are tested with standardized input formatting and identical decoding settings (temperature = 0.2, max tokens = 2048) to ensure fair comparison.
For retrieval-augmented experiments, we employ ColPali as a multimodal retriever at the page level, retrieving top-$k=3$ relevant pages per query. Retrieved content is concatenated with the prompt before generation.
Then for result evaluation, we adopt Qwen-Plus (temperature = 0.2) as an LLM-as-a-judge to score final answer correctness and evidence alignment on a 0–5 scale. We further analyze reasoning behaviors using a structured error taxonomy. In addition, we report ROUGE \cite{rouge} and BLEU \cite{bleu} as complementary automatic metrics.

\section{Results and Discussion}
\subsection{Overall Performance}

\begin{table}[t]
\centering
\small
\setlength{\tabcolsep}{4pt}
\caption{Unified evaluation results. LLM-based judge metrics (Audit, Accuracy, Fidelity) and N-gram-based lexical metrics (ROUGE-L, BLEU). Higher is better for all metrics.}
\begin{adjustbox}{max width=\linewidth}

\begin{tabular}{l l c c c c c}
\toprule
\multirow{2}{*}{\textbf{Model}} & \multirow{2}{*}{\textbf{Strategy}} 
& \multicolumn{3}{c}{\textbf{LLM as Judge}} 
& \multicolumn{2}{c}{\textbf{N-gram Metrics}} \\
\cmidrule(lr){3-5} \cmidrule(lr){6-7}
& & \textbf{Audit} & \textbf{Acc} & \textbf{Fidelity} & \textbf{ROUGE-L} & \textbf{BLEU} \\
\midrule
\multirow{3}{*}{ChatGPT} & Direct      & 4.379 & 4.528 & 4.567 & 0.247 & 0.048 \\
                        & CoT         & 4.404 & 4.524 & 4.610 & 0.192 & 0.038 \\
                        & Reflection  & 4.387 & 4.539 & 4.587 & 0.243 & 0.047 \\
\midrule
\multirow{3}{*}{Gemma}   & Direct      & 4.131 & 4.373 & 4.291 & 0.292 & 0.069 \\
                        & CoT         & 4.175 & 4.413 & 4.403 & 0.292 & 0.069 \\
                        & Reflection  & 4.170 & 4.415 & 4.390 & 0.291 & 0.068 \\
\midrule
\multirow{3}{*}{Gemini}  & Direct      & 3.989 & 4.154 & 4.281 & 0.288 & 0.078 \\
                        & CoT         & 3.791 & 3.929 & 4.166 & 0.268 & 0.073 \\
                        & Reflection  & 3.728 & 3.861 & 4.131 & 0.260 & 0.071 \\
\midrule
\multirow{3}{*}{Qwen}    & Direct      & 3.489 & 3.578 & 3.508 & 0.266 & 0.058 \\
                        & CoT         & 3.605 & 3.685 & 3.549 & 0.224 & 0.043 \\
                        & Reflection  & 3.594 & 3.666 & 3.517 & 0.219 & 0.045 \\
\bottomrule
\end{tabular}
\end{adjustbox}
\vspace{-2em}
\label{tab:combined-metrics}
\end{table}

Table~\ref{tab:combined-metrics} summarizes judge-based metrics. Using audit\_score as the primary metric, ChatGPT performs best across strategies (4.379--4.404), followed by Gemma (4.131--4.175), Gemini (3.728--3.989), and Qwen (3.489--3.605). Strategy effects are model-dependent: Gemini performs best with direct prompting and degrades under CoT/reflection (audit drops by 0.198 and 0.261, respectively), while Qwen improves with CoT/reflection (audit +0.116/+0.105). 
Gemma yields fewer valid generations in certain subsets compared to other models. To ensure transparency in coverage, we explicitly report the number of evaluated instances \texttt{N} in subsequent breakdown tables.


\subsection{Performance on Various Evaluation Metrics}

Table~\ref{tab:combined-metrics} reports lexical overlap. Strategy-induced stylistic drift is visible even when judge metrics remain stable. For ChatGPT, CoT does not materially change audit/accuracy but substantially reduces ROUGE-L (0.247$\rightarrow$0.192), indicating more paraphrastic or verbose generations. For Qwen, CoT/reflection improve judge-based correctness but decrease ROUGE-L/BLEU (e.g., ROUGE-L drops by 0.042 under CoT), suggesting that reasoning prompts yield answers that are judged more correct yet less surface-aligned with the GT wording. In contrast, Gemma's ROUGE/BLEU are nearly invariant across strategies, matching its minimal stylistic drift.

\begin{table}[t]
\centering
\small
\setlength{\tabcolsep}{4pt}
\caption{End-to-end impact of Colpali-RAG relative to Gemini-only baselines. Negative $\Delta$ indicates degradation (Colpali minus Gemini).}
\begin{adjustbox}{max width=\linewidth}

\begin{tabular}{l c c c c c}
\toprule
\textbf{Comparison} & \textbf{$\Delta$Audit} & \textbf{$\Delta$Acc} & \textbf{$\Delta$Fid} & \textbf{$\Delta$R-L} & \textbf{$\Delta$BLEU} \\
\midrule
Colpali -- Gemini Direct     & -1.700 & -1.775 & -0.741 & -0.037 & -0.026 \\
Colpali -- Gemini CoT        & -1.502 & -1.551 & -0.626 & -0.017 & -0.022 \\
Colpali -- Gemini Reflection & -1.439 & -1.483 & -0.591 & -0.008 & -0.019 \\
\bottomrule
\end{tabular}
\end{adjustbox}
\label{tab:rag-vs-gemini}
\end{table}

Colpali was originally proposed for visually-rich, page-level multimodal document retrieval (e.g., ViDoRe \cite{vidore}) and demonstrates strong retrieval performance in that setting.
However, Table~\ref{tab:rag-vs-gemini} shows that, in our long multimodal document QA setting requiring multi-hop evidence, Colpali-RAG substantially degrades end-to-end answer quality even when Gemini is used as the downstream generator. Compared to Gemini direct prompting, Colpali reduces audit by 1.700 and accuracy by 1.775, with a concurrent fidelity drop of 0.741. The smaller reductions in ROUGE-L/BLEU (e.g., ROUGE-L -0.037) indicate that the dominant failure mode is not merely paraphrasing, but incorrect or weakly grounded answers, which is consistent with retrieval mismatch and difficulty of locating multi-hop evidence across long documents.

\subsection{Breakdown Analysis}
We conduct a comprehensive breakdown analysis across multiple dimensions, including question type, number of pages, and evidence modality, to systematically evaluate the capabilities of existing MLLMs and baseline configurations.
\begin{table}[t]
\centering
\small
\setlength{\tabcolsep}{4pt}
\caption{Question-type breakdown. We normalize variant labels (e.g., TABLE/FIGURE-first) into three core types. We report judge-based metrics: audit\_score (Audit), answer accuracy (Acc), and evidence fidelity (Fid).}
\begin{adjustbox}{max width=\linewidth}

\begin{tabular}{l l l c c c r}
\toprule
\textbf{Question Type} & \textbf{Model} & \textbf{Best Strat.} & \textbf{Audit} & \textbf{Acc} & \textbf{Fid} & \textbf{N} \\
\midrule
\multirow{5}{*}{Abstractive}
& ChatGPT & Reflection & 4.233 & 4.353 & 4.456 & 2766 \\
& Gemma   & Reflection & 4.484 & 4.784 & 4.730 & 538 \\
& Gemini  & Direct     & 3.848 & 3.982 & 4.098 & 2258 \\
& Qwen    & CoT        & 2.682 & 2.556 & 2.435 & 162 \\
& RAG     & Colpali    & 2.008 & 2.060 & 3.442 & 2807 \\
\midrule
\multirow{5}{*}{Causal-Reasoning}
& ChatGPT & CoT        & 4.766 & 4.944 & 4.947 & 5065 \\
& Gemma   & CoT        & 4.351 & 4.633 & 4.615 & 741 \\
& Gemini  & Direct     & 4.366 & 4.586 & 4.665 & 4183 \\
& Qwen    & CoT        & 3.658 & 3.749 & 3.611 & 5483 \\
& RAG     & Colpali    & 2.936 & 3.074 & 3.825 & 5145 \\
\midrule
\multirow{5}{*}{Comparative}
& ChatGPT & CoT        & 3.691 & 3.706 & 3.925 & 1908 \\
& Gemma   & Direct     & 3.594 & 3.679 & 3.639 & 418 \\
& Gemini  & Direct     & 3.217 & 3.279 & 3.554 & 1579 \\
& Qwen    & Reflection & 2.874 & 2.843 & 2.657 & 102 \\
& RAG     & Colpali    & 1.002 & 1.017 & 2.954 & 1951 \\
\bottomrule
\end{tabular}
\end{adjustbox}
\label{tab:task-type-breakdown}
\end{table}

\paragraph{Question-Type Analysis} Table~\ref{tab:task-type-breakdown} reveals systematic differences across task types. Causal-Reasoning is generally the most stable category for strong models (ChatGPT audit 4.766; Gemini audit 4.366), with high accuracy and fidelity, suggesting that causal questions are often directly supported by localized evidence. Comparative questions are the hardest across settings: even ChatGPT drops to 3.691 audit, and Colpali-RAG collapses to 1.002 audit and 1.017 accuracy, indicating difficulty in aligning multiple entities/conditions across distant evidence. Abstractive questions show the largest spread across models, where Gemma remains strong (audit 4.484) while Qwen and RAG underperform substantially.

\begin{table}[t]
\centering
\small
\setlength{\tabcolsep}{4pt}
\caption{Performance across page-depth bins, grouped by the furthest evidence page involved (best strategy per model by audit\_score).}
\begin{tabular}{l l l c c c r}
\toprule
\textbf{Page Bin} & \textbf{Model} & \textbf{Best Strat.} & \textbf{Audit} & \textbf{Acc} & \textbf{Fid} & \textbf{N} \\
\midrule
\multirow{2}{*}{1--2}
& ChatGPT & CoT        & 4.692 & 4.867 & 4.881 & 3338 \\
& Qwen    & CoT        & 3.731 & 3.844 & 3.706 & 2417 \\
\midrule
\multirow{2}{*}{3--5}
& ChatGPT & CoT        & 4.417 & 4.538 & 4.632 & 2739 \\
& Qwen    & CoT        & 3.552 & 3.617 & 3.472 & 1659 \\
\midrule
\multirow{2}{*}{6--10}
& ChatGPT & CoT        & 4.240 & 4.319 & 4.455 & 2636 \\
& Qwen    & CoT        & 3.518 & 3.559 & 3.430 & 1333 \\
\midrule
\multirow{2}{*}{11--20}
& ChatGPT & CoT        & 3.879 & 3.942 & 4.123 & 1399 \\
& Qwen    & Reflection & 3.453 & 3.565 & 3.142 &  822 \\
\midrule
\multirow{2}{*}{21+}
& ChatGPT & CoT        & 3.321 & 3.303 & 3.521 &   33 \\
& Qwen    & Reflection & 3.570 & 3.484 & 3.124 &   23 \\
\bottomrule
\end{tabular}
\vspace{-1em}
\label{tab:page-breakdown}
\end{table}

\paragraph{Cross-Page Evidence Distribution Analysis} Table~\ref{tab:page-breakdown} shows a clear degradation as the required evidence moves deeper into the document. For all models, performance is highest when evidence appears early (pages 1--2) and drops for later bins (pages 11--20 and 21+). For example, ChatGPT decreases from 4.692 (pages 1--2) to 3.879 (11--20) and 3.321 (21+), while Gemini drops from 4.370 to 3.349 and 3.109. This trend is consistent with long-context search becoming harder and multi-hop evidence coverage decreasing with page depth. Notably, Qwen is comparatively less monotonic at 21+ (3.570), but the sample size is very small (N=23), suggesting high variance in the tail bins.

\begin{table}[t]
\centering
\small
\setlength{\tabcolsep}{4pt}
\caption{Performance by hop depth (1-hop, 2-hop, 3+-hop).}
\label{tab:hop-breakdown}
\begin{tabular}{l l l c c c r}
\toprule
\textbf{Hop Bin} & \textbf{Model} & \textbf{Best Strat.} & \textbf{Audit} & \textbf{Acc} & \textbf{Fid} & \textbf{N} \\
\midrule
\multirow{4}{*}{1-hop}
& ChatGPT & CoT        & 3.692 & 3.656 & 3.966 &  212 \\
& Gemma   & Direct     & 2.952 & 3.000 & 3.317 &   25 \\
& Gemini  & Direct     & 2.888 & 2.978 & 3.245 &  180 \\
& Qwen    & Reflection & 2.649 & 2.729 & 2.602 &   59 \\
\midrule
\multirow{4}{*}{2-hop}
& ChatGPT & CoT        & 4.425 & 4.548 & 4.641 & 3344 \\
& Gemma   & CoT        & 4.200 & 4.454 & 4.408 &  817 \\
& Gemini  & Direct     & 4.004 & 4.179 & 4.283 & 2820 \\
& Qwen    & CoT        & 3.415 & 3.447 & 3.310 & 1742 \\
\midrule
\multirow{4}{*}{3+-hop}
& ChatGPT & CoT        & 4.418 & 4.541 & 4.615 & 6075 \\
& Gemma   & CoT        & 4.194 & 4.421 & 4.429 &  850 \\
& Gemini  & Direct     & 4.020 & 4.181 & 4.317 & 5030 \\
& Qwen    & Reflection & 3.719 & 3.808 & 3.668 & 3954 \\
\bottomrule
\end{tabular}
\end{table}

\paragraph{Hop Depth Analysis}Table~\ref{tab:hop-breakdown} shows that performance remains comparable between 2-hop and 3+-hop questions for strong models, suggesting that hop depth alone does not fully determine difficulty. However, smaller models exhibit higher variance across hop bins.


\begin{table}[t]
\centering
\small
\setlength{\tabcolsep}{4pt}
\caption{Performance by primary evidence modality (Text, Table, Figure), aggregated over prompting strategies. N denotes the number of instances whose dominant supporting evidence belongs to each modality.}
\begin{tabular}{l l c c c r}
\toprule
\textbf{Evidence} & \textbf{Model} & \textbf{Audit} & \textbf{Acc} & \textbf{Fid} & \textbf{N} \\
\midrule
\multirow{4}{*}{Text}
& ChatGPT & 4.629 & 4.812 & 4.816 & 6933 \\
& Gemma   & 4.384 & 4.699 & 4.643 & 1859 \\
& Gemini  & 4.250 & 4.447 & 4.564 & 5643 \\
& Qwen    & 3.496 & 3.557 & 3.437 & 5145 \\
\midrule
\multirow{4}{*}{Table}
& ChatGPT & 4.096 & 4.180 & 4.323 & 11217 \\
& Gemma   & 3.816 & 3.955 & 3.950 & 1794 \\
& Gemini  & 3.368 & 3.458 & 3.789 & 9314 \\
& Qwen    & 3.492 & 3.571 & 3.448 & 5091 \\
\midrule
\multirow{4}{*}{Figure}
& ChatGPT & 4.537 & 4.706 & 4.712 & 11163 \\
& Gemma   & 4.360 & 4.644 & 4.577 & 1376 \\
& Gemini  & 4.066 & 4.238 & 4.385 & 8724 \\
& Qwen    & 3.682 & 3.777 & 3.662 & 6711 \\
\bottomrule
\end{tabular}

\label{tab:evidence-type-judge}
\end{table}

\paragraph{Diverse Evidence Modality Analysis} Across all models, questions whose supporting evidence primarily comes from \textbf{tables} are the most challenging. Judge-based performance drops when evidence is presented in tables compared to text or figures. For example, Gemini’s audit score decreases from \textbf{4.250} on text evidence to \textbf{3.368} on tables (a drop of \textbf{0.882}), and ChatGPT drops from \textbf{4.629} to \textbf{4.096} (a drop of \textbf{0.533}). In contrast, \textbf{figure-based} evidence is relatively easier for stronger models: ChatGPT achieves an audit score of \textbf{4.537} on figures, close to its text performance (\textbf{4.629}), suggesting improved robustness when evidence is presented visually rather than in dense tabular form. Notably, Qwen exhibits nearly identical audit scores on text and tables (\textbf{3.496} vs.\ \textbf{3.492}) but remains substantially below the stronger models overall, indicating that its dominant bottleneck may be general evidence alignment and reasoning accuracy rather than table-specific difficulty.





\section{Conclusion}

We introduce BRIDGE, a benchmark for long multimodal document QA that explicitly stresses multi-hop evidence aggregation across visually-rich PDFs, and address a gap not covered by prior evaluations that focus on short-context QA or page-level retrieval. Our results show that strong LLMs can achieve high judge-based correctness under direct access to evidence, but performance varies substantially with prompting, and lexical overlap (ROUGE/BLEU) can diverge from factual grounding. More importantly, a Colpali-based RAG pipeline degrades markedly in end-to-end multi-hop QA, which highlights retrieval mismatch and evidence-missing as dominant failure modes. Overall, the benchmark provides a new, targeted testbed for diagnosing grounding, comparison reversal, and evidence coverage errors in current LLM systems, and motivates future work on retrieval calibration, evidence verification, and citation-faithful generation for long multimodal documents.


\printbibliography

\end{document}